%% file: neurips_2026.tex
\documentclass{article}

\usepackage[main, preprint]{neurips_2026}

\usepackage[utf8]{inputenc} 
\usepackage[T1]{fontenc}    
\usepackage{hyperref}       
\usepackage{url}            
\usepackage{booktabs}       
\usepackage{amsfonts}       
\usepackage{nicefrac}       
\usepackage{xcolor}         

\usepackage{graphicx}
\usepackage{amsmath}
\usepackage{enumitem}
\title{KV Packet: Recomputation-Free Context-Independent KV Caching for LLMs}

\author{
  Chuangtao Chen \\
  Technical University of Munich \\
  Munich, Germany \\
  \texttt{chuangtao.chen@tum.de} \\
  \And
  Grace Li Zhang \\
  Technical University of Darmstadt \\
  Darmstadt, Germany \\
  \texttt{grace.zhang@tu-darmstadt.de} \\
  \And
  Xunzhao Yin \\
  Zhejiang University \\
  Hangzhou, China \\
  \texttt{xzyin1@zju.edu.cn} \\
  \And
  Cheng Zhuo \\
  Zhejiang University \\
  Hangzhou, China \\
  \texttt{czhuo@zju.edu.cn} \\
  \And
  Bing Li \\
  Technische Universität Ilmenau \\
  Ilmenau, Germany \\
  \texttt{bing.li@tu-ilmenau.de} \\
  \And
  Ulf Schlichtmann \\
  Technical University of Munich \\
  Munich, Germany \\
  \texttt{ulf.schlichtmann@tum.de} \\
}
\begin{document}
\maketitle
\input{01_abstract}
\input{02_introduction}
\input{03_background}
\input{04_method}

\input{05_experiment}

\input{06_conclusion}

\begin{ack}
This work is funded by the Deutsche Forschungsgemeinschaft (DFG, German Research Foundation) – Project-ID 504518248 and in part by the European Union - European Research Council (ERC) Starting Grant -Project-ID 101219243. Views and opinions expressed are however those of the author(s) only and do not necessarily reflect those of the European Union or the European Research Council. Neither the European Union nor the granting authority can be held responsible for them.
\end{ack}

\bibliography{iclr2025_conference}
\bibliographystyle{iclr2025_conference}

\begingroup
\setlength{\textfloatsep}{8pt plus 2pt minus 2pt}
\setlength{\intextsep}{8pt plus 2pt minus 2pt}
\setlength{\floatsep}{8pt plus 2pt minus 2pt}
\input{07_appendix}

\clearpage
\endgroup
\end{document}

%% file: 01_abstract.tex
\begin{abstract}
Large Language Models (LLMs) rely heavily on Key-Value (KV) caching to minimize inference latency. However, standard KV caches are context-dependent: reusing a cached document in a new context requires recomputing KV states to account for shifts in attention distribution. Existing solutions such as CacheBlend, EPIC, and SAM-KV mitigate this issue by selectively recomputing a subset of tokens; however, they still incur non-negligible computational overhead (FLOPs) and increased Time-to-First-Token (TTFT) latency. In this paper, we propose KV Packet, a recomputation-free cache reuse framework that treats cached documents as immutable ``packets'' wrapped in light-weight trainable soft-token adapters, which are trained via self-supervised distillation to bridge context discontinuities. Experiments on Llama-3.1 and Qwen-3 demonstrate that the proposed KV Packet method achieves near-zero FLOPs and lower TTFT than recomputation-based baselines, while retaining F1 scores comparable to those of the full recomputation baseline. Code and experimental results are available at \url{https://github.com/ChuangtaoChen-TUM/KVPacket}.
\end{abstract}

%% file: 02_introduction.tex
\section{Introduction}\label{sec:introduction}

Large Language Models (LLMs) \citep{llama3, jiang2023mistral7b, liu2024deepseek, wang2024} are increasingly deployed in Retrieval-Augmented Generation (RAG) systems, where a user query is answered by first retrieving a set of relevant documents from a knowledge base and then prompting the LLM with those documents as context. In production settings, the same documents are often retrieved repeatedly across many different user queries. A natural optimization is therefore to \textit{precompute} and \textit{cache} the Key-Value (KV) states of these documents offline, so that at inference time, the system can simply load the cached values rather than reprocessing the raw text from scratch. This avoids the expensive `prefill' phase \citep{vllm, flashattention, lin2025}, where input tokens are processed to generate the KV cache, and can substantially reduce Time-to-First-Token (TTFT) latency.

However, this optimization breaks down in practice due to the \textit{contextual dependence} of KV caches. Because the KV states for each token are conditioned on its entire preceding context, a cache precomputed for a document in isolation is no longer valid when that document is placed alongside other retrieved chunks at inference time. Naively concatenating independently precomputed KV caches causes catastrophic performance degradation, since each cache encodes attention distributions that are oblivious to the surrounding documents. Consequently, the attention computation for each document is performed over an incomplete prefix, failing to capture the cross-document dependencies that would arise during full inference. Existing solutions fall into two categories: fine-tuning-based adaptation (KVLink \citep{kvlink}, Block-Attention \citep{block_attention}, CacheClip \citep{cacheclip}), which risks degrading the base model's general capabilities, and inference-time selective recomputation (CacheBlend \citep{cacheblend}, EPIC \citep{epic}, A3 \citep{a3}), which reintroduces TTFT latency by performing partial forward passes at serving time. Neither approach achieves truly recomputation-free, plug-and-play cache reuse.

To address these limitations, we propose KV Packet, a framework that enables context-independent KV caching without model fine-tuning or inference-time recomputation. By wrapping document blocks with learned shared adapters, KV Packet ensures that precomputed caches are intrinsically compatible for direct concatenation. This decoupling enables seamless composition of retrieved documents at serving time with zero additional computational overhead.

The key contributions of this paper are as follows:
\begin{enumerate}[leftmargin=1.2em, itemsep=0pt, topsep=2pt]
    \item \textbf{Zero-Recomputation Architecture}: We introduce the KV Packet abstraction, which wraps frozen document caches with trainable soft-token adapters. This design eliminates the need for model parameter modifications and expensive cache recomputation during inference.
    \item \textbf{Self-Supervised Alignment}: We propose a self-distillation training objective for the trainable fillers that aligns disjointed KV packets with full-attention representations. This approach requires no human-labeled data and restores contextual connectivity without modifying the base model's parameters.
    \item \textbf{Efficiency and Performance}: We demonstrate that KV Packet reduces computational overhead (FLOPs) by approximately 4 orders of magnitude compared to recomputation-based methods like CacheBlend and EPIC, while maintaining competitive performance on various tasks.
    \item \textbf{Compatibility with KV Compression}: As a recomputation-free architecture, KV Packet natively supports existing KV compression techniques seamlessly, while recomputation-based methods struggle with unstructured compression with varying token preservation across layers.
\end{enumerate}

The remainder of this paper is organized as follows: Section~\ref{sec:background} reviews KV caching mechanics and related work. Section~\ref{sec:methods} details the KV Packet architecture and self-supervised distillation objective. Section~\ref{sec:experiments} evaluates performance, latency, and computational efficiency against recomputation-based baselines, including compatibility experiments and ablation studies. Section~\ref{sec:conclusion} summarizes our findings and discusses limitations and future directions.

%% file: 03_background.tex
\section{Background and Related Work}\label{sec:background}
\begin{figure}
    \centering
    \includegraphics[width=\linewidth]{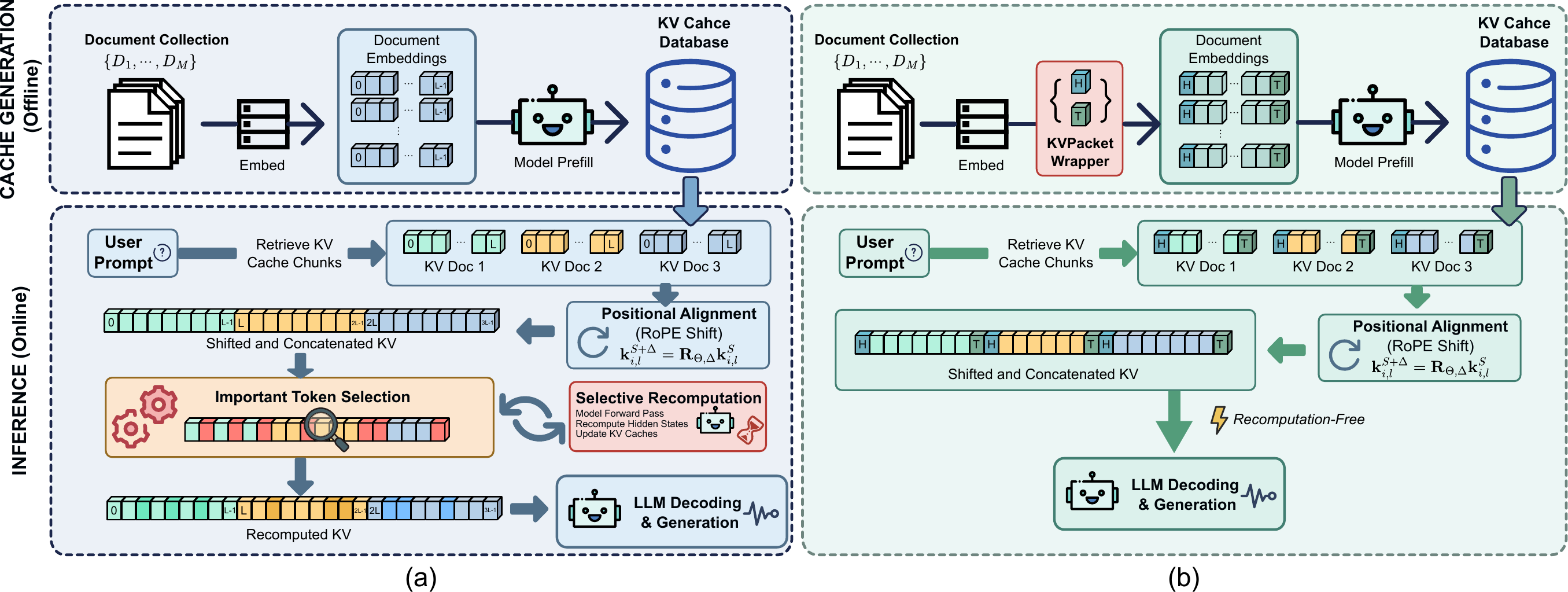}
    \caption{Comparison of KV cache reuse architectures. (a) Recomputation-based approaches require inference-time algorithms to select and recompute important tokens to repair contextual staleness; (b) the proposed KV Packet approach wraps documents with global adapters.}
    \label{fig:pipeline}
\end{figure}

\subsection{Key-Value Caching and Positional Encoding}
Efficient LLM inference relies on a two-phase execution: a \textit{prefill} stage that processes the input prompt in parallel to produce the Key-Value (KV) states, and an auto-regressive \textit{decoding} stage that reuses cached states to avoid redundant computation \citep{transformer}. Industrial providers exploit this by caching recurring contexts across sessions \citep{cheng2025lmcache, promptcache, openai2024promptcaching, anthropic2024promptcaching, google2026contextcaching}.

However, KV caches are fragile due to \textbf{prefix dependency} and \textbf{positional dependency}: KV states are conditioned on both the full preceding context and absolute position indices, therefore, a precomputed cache is only valid at an identical position with an identical prefix. Even shifting a document by one token invalidates the entire cache. This sensitivity prevents the modular reuse of document caches in dynamic environments like multi-document RAG, necessitating expensive full recomputation for every unique configuration.

The positional dependency admits an efficient closed-form solution. Modern LLMs (\textit{e.g.}, Llama~3 \citep{llama3}, Mistral \citep{jiang2023mistral7b}, Qwen~3 \citep{yang2025qwen3technicalreport}) adopt Rotary Positional Embeddings (RoPE) \citep{rope}, which encode position by rotating Query and Key vectors by a position-dependent angle. Since the attention score between any two tokens depends only on their relative displacement, a precomputed Key state $\mathbf{k}_{i}^{S}$ can be efficiently realigned to a new position $s+\Delta$ via a single rotation \citep{rope, kvlink}:
\begin{equation}\label{eq:rope_shift}
    \mathbf{k}_{i}^{S+\Delta}=\mathbf{R}_{\Theta, \Delta}\mathbf{k}_{i}^{S}, \quad \forall \Delta \in \mathbb{Z},
\end{equation}
where $\mathbf{R}_{\Theta, m}$ is the RoPE rotation matrix encoding offset $\Delta$. This operation is element-wise and computationally negligible compared to a full forward pass, effectively resolving positional dependency at near-zero cost.

\begin{figure}[t]
    \centering
    \includegraphics[width=.9\linewidth]{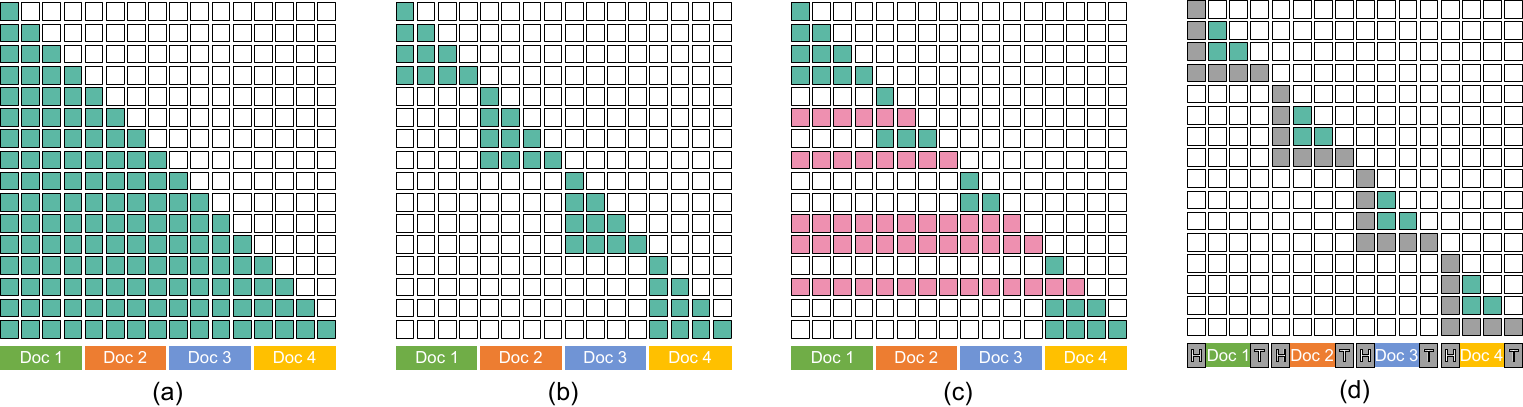}
    \caption{Comparison of mathematically equivalent attention maps. (a) Full Recomputation. (b) Naive Concatenation. (c) Selective Recomputation (dashed blocks for recomputed tokens). (d) KV Packet: Attention with lightweight adapters (\textbf{H}eader and \textbf{T}railer).}
    \label{fig:attn_map}
\end{figure}

However, positional alignment alone is insufficient. The model still suffers from catastrophic performance degradation due to context dependency. Fig.~\ref{fig:attn_map} illustrates this phenomenon through the lens of attention maps. In an ideal full-prefill scenario (Fig.~\ref{fig:attn_map}(a)), the model computes hidden states within a continuous, full attention map where every token has visibility of the entire preceding stream. In contrast, naive concatenation (Fig.~\ref{fig:attn_map}(b)) creates isolated attention blocks. Because each document's KV cache was generated independently, its internal representations were computed using an incomplete attention map that lacks the semantic influence of the global prefix.

\subsection{Context-Independent KV Caching}
\textbf{Model-Modification Approaches}: To address the challenges of contextual dependency in cache reuse without the cost of a full prefill, several recent works have proposed extending context capabilities by updating model parameters or introducing auxiliary networks. KVLink \citep{kvlink}, for instance, introduces trainable link tokens coupled with explicit fine-tuning of the base model, while Block-Attention \citep{block_attention} modifies the attention mechanism itself to accommodate block-wise processing. Similarly, CacheClip \citep{cacheclip} relies on fine-tuning another language model to guide token selection.

Despite their effectiveness, we distinguish these fine-tuning and model-modification approaches from our work due to several practical drawbacks. Fine-tuning the base models or auxiliary supervisors requires massive computational resources. Furthermore, methods like CacheClip \citep{cacheclip} increase deployment complexity by requiring the simultaneous serving of multiple models, which significantly raises memory pressure. Most importantly, modifying the base model’s weights can lead to catastrophic forgetting, potentially degrading the model’s performance on general tasks.

\textbf{Recomputation-Based Approaches}: A complementary line of work addresses cache staleness through \textit{selective recomputation}: recomputing only a critical subset of tokens at inference time to repair the missing prefix context, as illustrated in Fig.~\ref{fig:attn_map}(c). CacheBlend \citep{cacheblend} identifies high-deviation tokens across layers; A3 \citep{a3} selects tokens by real-time query-document attention scores; EPIC \citep{epic} recomputes anchor tokens at document boundaries; and SAM-KV \citep{samkv} applies hierarchical compression for multi-context scenarios. Despite their efficiency gains over full prefill, these methods share the following drawbacks:
\begin{itemize}[leftmargin=1.2em, itemsep=0pt, topsep=2pt]
\item \textbf{Increased TTFT}: These methods must perform additional forward passes to repair the KV cache before generation, which can be further exacerbated by the complex token selection algorithms.
\item \textbf{Computational overhead}: While recomputing only a subset of tokens is lighter than a full prefill, the cumulative FLOPs of these methods remain substantial.
\item \textbf{Engineering burden}: These strategies are inherently invasive, requiring deep integration with the model's internal attention mechanisms and forward-pass logic. Consequently, adapting these methods to different model architectures requires extensive code modifications.
\end{itemize}

\subsection{Soft-token Adapter Methods}
Another line of work conceptually related to our method is parameter-efficient adaptation via learnable soft tokens prepended to the input. Prefix-tuning \citep{li-liang-2021-prefix} and prompt tuning \citep{lester-etal-2021-power} optimize continuous vectors that act as virtual prompt tokens while keeping the backbone LLM frozen, enabling efficient task conditioning without updating model weights. KV Packet similarly employs a small set of trainable tokens; however, instead of task adaptation from scratch, our goal is \emph{cache composition}: training Header/Trailer tokens that absorb boundary artifacts and reconcile disjoint KV caches. Therefore, independently prefetched documents can be stitched together without inference-time recomputation.

%% file: 04_method.tex
\section{Methods}
\label{sec:methods}

When concatenating precomputed KV caches with RoPE realignment, the cached keys for each document remain contextually stale because they were computed without regard for the preceding documents. Prior recomputation-based methods address this by selectively recomputing a subset of tokens to restore cross-document attention. We hypothesize that a significant contributor to performance degradation in naive concatenation is not solely the absence of cross-document attention, but rather \textbf{boundary artifacts}, specifically, the disruption of attention sinks and abrupt shifts in token distributions at block boundaries, which interfere with the model's attention mechanism. This is partly motivated by the observation that retrieved documents in typical RAG settings are often semantically independent units, suggesting that the model's primary challenge at inference time is integrating information across structurally discontinuous boundaries rather than resolving deep semantic dependencies between documents. This perspective motivates wrapping each document cache with lightweight learned adapters that act as smooth delimiters at block boundaries, enabling the query to integrate information from independent cached packets without being disrupted by structural discontinuities.

\subsection{Formulation of KV Packet}

\textbf{KV Packet Abstraction}: We define a KV Packet $\mathcal{P}(D; \phi)$ as the KV cache of a frozen document $D$ encapsulated with learnable boundary adapters $\phi = \{\mathbf{H}, \mathbf{T}\}$, where the header $\mathbf{H} = [\mathbf{h}_1, \dots, \mathbf{h}_{N_h}] \in \mathbb{R}^{N_h \times d}$ and the trailer $\mathbf{T} =[\mathbf{t}_1, \dots, \mathbf{t}_{N_t}] \in \mathbb{R}^{N_t \times d}$ are sequences of $N_h$ and $N_t$ continuous vectors, respectively. Denoting $\mathbf{e}_i$ as the embedding of the $i$-th document token, the packet representation for a single document is:

\begin{equation}\label{eq:packet_const}
\mathcal{P}(D;\phi) = [\mathbf{h}_1, \dots, \mathbf{h}_{N_h}, \mathbf{e}_1, \dots, \mathbf{e}_L, \mathbf{t}_1, \dots, \mathbf{t}_{N_t}]
\end{equation}

At inference time, each document is independently wrapped into a packet according to Eq.~\eqref{eq:packet_const} and its KV cache is precomputed offline. The resulting caches are then directly concatenated at serving time without any recomputation. The adapters $\phi$ are trained to act as shared smooth delimiters at block boundaries, such that the concatenated packets yield output quality comparable to full-attention inference, without requiring cross-document attention.

Fig.~\ref{fig:pipeline} shows the pipelines of recompute-based KV cache reuse methods and the proposed KV Packet approach. Recomputation-based methods (Fig.~\ref{fig:pipeline}(a)) require an online token selection and partial forward pass step before generation, incurring significant FLOPs and latency. In comparison, as shown in Fig.~\ref{fig:pipeline}(b), KV Packet confines all adapter-related work to the offline cache generation stage: document embeddings are wrapped with $\mathbf{H}$ and $\mathbf{T}$ at the embedding level prior to KV generation, requiring no architecture-specific implementation. At serving time, the recomputation step is eliminated entirely; only lightweight positional alignment and KV concatenation remain, effectively constructing global context without re-executing attention or FFN layers over document tokens. This results in an equivalent attention map illustrated in Fig.~\ref{fig:attn_map}(d) where the Header and Trailer adapters serve as structural boundaries between document blocks, enabling full cross-document visibility for the query without recomputing any document tokens.

Beyond the serving-time efficiency gains described above, the parameters $\phi$ are global and document-independent: learned once on a representative text distribution and shared across all documents. New documents are wrapped with pre-trained adapters offline, incurring a constant $O(1)$ cost of a few kilobytes for the $N_h+N_t$ vectors, regardless of the size of the retrieval corpus. Furthermore, the total adapter length is very small relative to the typical document length; the additional KV cache storage and computation introduced by the adapters is also negligible.

\subsection{Training via Self-Supervised Distillation}

To enable the KV packets to function cohesively, we propose a self-supervised knowledge distillation framework to optimize the adapters $\phi = \{\mathbf{H}, \mathbf{T}\}$ to mimic the behavior of a model with full-context visibility, where the model $\mathcal{M}$ serves as its own teacher. The framework consists of two passes: a \textbf{reference pass}, in which the model processes the full context with standard causal attention to produce the target distribution; and a \textbf{student pass}, in which the model processes the same context using the KV Packet architecture to produce the approximated distribution. The adapters are then optimized by minimizing the divergence between the two distributions.

\subsubsection{Reference Generation and Student Pass}
\begin{enumerate}[leftmargin=1.2em, itemsep=0pt, topsep=2pt]
    \item \textbf{Input Construction}: We randomly sample a set of documents $\{D_1, \dots, D_M\}$ and a corresponding query $Q$ from the training corpus. These are concatenated to form the context: $X_{context} = \left[D_1, \cdots, D_M, Q\right]$.
    \item \textbf{Auto-regressive Generation}: The model processes $X_{context}$ using standard causal attention, generating a continuation sequence $G=\left[g_1,\cdots,g_T\right]$ auto-regressively. The resulting token distribution $\mathbf{P}_{teacher} \in \mathbb{R}^{|G| \times |\mathcal{V}|}$ serves as the golden reference.
    \item \textbf{Packet Construction}: Each document $D_i$ is wrapped to form $\mathcal{P}(D_i;\phi)$ according to Eq.~\eqref{eq:packet_const}. Then, we construct the KV caches with adapters for all $\mathcal{P}(D_i;\phi)$ independently.
    \item \textbf{Forward with KV Packets}: KV caches with adapters are realigned via Eq.~\eqref{eq:rope_shift} and then concatenated into a single KV cache. A forward pass is performed on $\left[Q, G\right]$ given the KV cache with causal attention, where $\left[Q, G\right]$ can attend to all previous tokens. The probability distribution for $G$ is recorded as $\mathbf{P}_{student} \in \mathbb{R}^{|G| \times |\mathcal{V}|}$.
    \item \textbf{Loss Calculation}: We minimize the Kullback-Leibler Divergence between the student's output distribution and the teacher's reference distribution: $$\mathcal{L} = \frac{1}{|G|} \sum_{t=1}^{|G|} D_{KL}\left(\mathbf{P}_{teacher}^{t} \parallel \mathbf{P}_{student}^{t} \right),$$ where $\textbf{P}^t$ represents the probability distribution to predict token $g_t$ in the generated sequence.
\end{enumerate}

Utilizing the model's own full-attention generation as a teacher signal offers several advantages over standard fine-tuning. Since only the small adapter tensors receive gradients while the base model $\mathcal{M}$ and all document embeddings remain frozen, memory and FLOPs are kept significantly below those of fine-tuning methods, and catastrophic forgetting is bypassed by design. Furthermore, no human-annotated data is required: adapters can be trained on diverse unlabeled corpora by treating random text segments as independent documents, and distilling the model's own distribution ensures that the adapters learn only the structural patterns needed to recover full-prefill information flow, rather than memorizing external facts.

%% file: 05_experiment.tex
\section{Experiments}
\label{sec:experiments}
\subsection{Experimental Setup}
We evaluate KV Packet on four datasets spanning two task types: simple information retrieval (Needle-in-a-Haystack \citep{niah_ruler}, Biography \citep{biographies}) and multi-step reasoning (HotpotQA \citep{yang2018hotpotqa}, MusiQue \citep{trivedi2021musique}). All experiments use a single NVIDIA A100 (80~GB) GPU with 8 adapter tokens ($N_h=N_t=8$), float32 adapters, and models in bfloat16. Adapters are trained for 30 epochs with AdamW, a linear decay schedule, and 256–512 training samples (batch size 64); learning rates are $5\times 10^{-4}$ for Biography, NIAH, and MusiQue, and $1\times 10^{-3}$ for HotpotQA. Cached KV tensors are stored in CPU memory and loaded to the GPU upon retrieval.

\begin{figure}[t]
    \centering
    \includegraphics[width=0.95\linewidth]{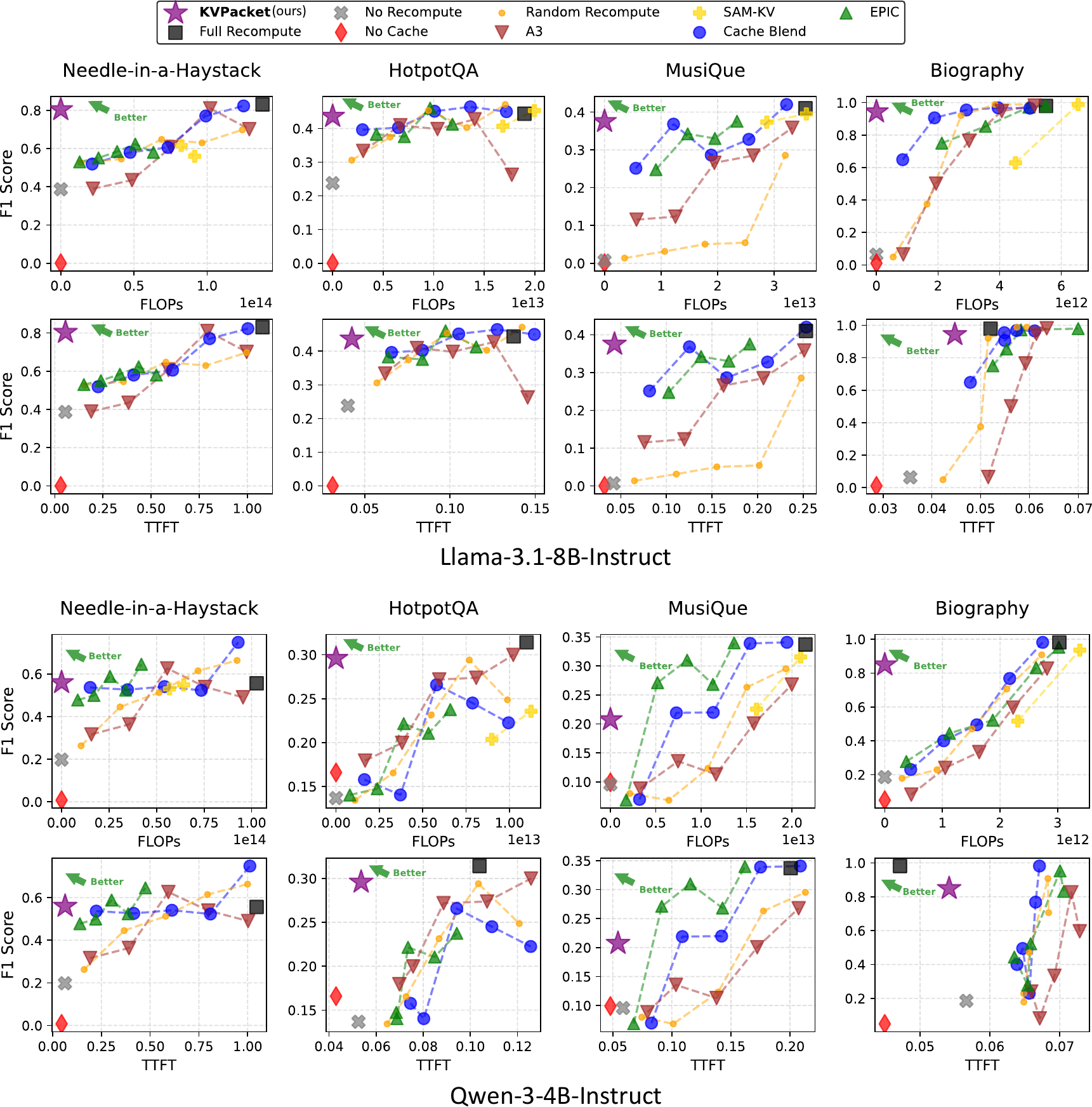}
    \caption{Evaluation results (F1 score, FLOPs, Time-to-First-Token in seconds) of Llama-3.1-8B / Qwen-3-4B on datasets: Needle-in-a-Haystack, Biography, HotpotQA, and MusiQue.}
    \label{fig:llama_result}
\end{figure}


We benchmarked against a comprehensive suite of baselines: Full Recompute (upper bound), No Recompute (lower bound, RoPE realignment only), No Cache (no documents, control baseline using only model knowledge), and the following recomputation-based methods: CacheBlend \citep{cacheblend}, A3 \citep{a3}, and Random recomputation with recomputation ratios: 10\%, 30\%, 50\%, 70\%, and 90\% of document tokens; EPIC \citep{epic} with 10, 30, 50, 70, and 90 recomputed tokens (and 100–500 for NIAH). SAM-KV \citep{samkv} evaluated under two configurations:  small: block/initial/local tokens = 8/8/16; large: 64/64/128. The configurations listed above correspond to the left-to-right order of data points in Fig.~\ref{fig:llama_result}.

\subsection{Efficiency Metrics}
\textbf{FLOPs} measure operations required to align and prepare retrieved caches before generation: for KV Packet, this is solely the RoPE position realignment; for recompute methods, it additionally includes the partial forward pass over selected tokens. \textbf{TTFT} is the end-to-end wall-clock latency from query receipt to first token, including cache transfer from CPU to GPU, RoPE shifts or recomputation, and query processing. Offline cache generation costs are excluded from inference metrics.

\subsection{Results and Analysis}
We present the performance of Llama-3.1-8B-Instruct and Qwen-3-4B-Instruct on the evaluated datasets in Fig.~\ref{fig:llama_result}.

\textbf{Generation Quality}: For most configurations, KV Packet achieves comparable F1 on information retrieval tasks (Needle-in-a-Haystack, Biography) and multi-step reasoning (HotpotQA, MusiQue), substantially outperforming No Recompute in all settings. Recomputation methods such as EPIC and CacheBlend struggle at low recompute ratios, particularly in long-context settings. For the Qwen model on the MusiQue dataset, the KV Packet method demonstrates a performance gap when compared with the Full Recompute baseline, but still maintains a favorable Pareto trade-off when considering the FLOPs and TTFT.

\textbf{Computational Efficiency}: With the recomputation-free nature of the proposed method, we achieve significant reductions compared with the Full Recompute baseline, with FLOPs by 5–6 orders of magnitude lower ($6.50\times10^{-6} \text{ to } 1.04\times 10^{-5}$), matching the No Recompute baseline. As shown in Fig.~\ref{fig:llama_result}, the KV Packet is located at the upper-left of the F1 score vs. FLOPs plot, achieving high generation quality while retaining low computational cost.

\textbf{Latency (TTFT)}: KV Packet consistently delivers one of the lowest Time-to-First-Token (TTFT) values, similar to No Recompute and only slightly higher than No Cache (which lacks the necessary context). For experiments on the Llama model, the KV Packet method achieves $1.36\times$ and $3.3\times$ speedup on the Biography and HotpotQA tasks, respectively. The TTFT reduction is especially significant in long-context settings: we observe a $19.45\times$ reduction in TTFT on Needle-in-a-Haystack and a $5.81\times$ reduction on MusiQue. As shown in the F1 score vs. TTFT plots in Fig.~\ref{fig:llama_result}, KV Packet occupies the upper-left region, indicating high generation quality at low latency.

Overall, the experimental results validate KV Packet as an effective alternative to recomputation-based strategies, recovering much of the same base model's Full Recompute accuracy on both extraction and reasoning tasks while eliminating the recomputation overhead entirely at inference time.
A direct comparison with KVLink \citep{kvlink} is reported in Appendix~\ref{app:kvlink}.

\subsection{Compatibility with KV Compression}

Recomputation-based methods face significant practical challenges when combined with modern KV compression: partial forward passes require complete, contiguous caches with known positional indices. However, compression algorithms produce unstructured caches that preserve different token subsets per layer, breaking RoPE realignment and selective attention assumptions. Furthermore, current compression algorithms are strictly optimized for auto-regressive generation rather than historical KV recomputation, and their efficacy and numerical stability in recomputation settings remain unverified. KV Packet natively bypasses these bottlenecks entirely by treating each document cache as an opaque unit and never re-evaluating compressed hidden states; it integrates seamlessly with off-the-shelf unstructured KV pruning.

\begin{figure}[t]
    \centering
    \includegraphics[width=\linewidth]{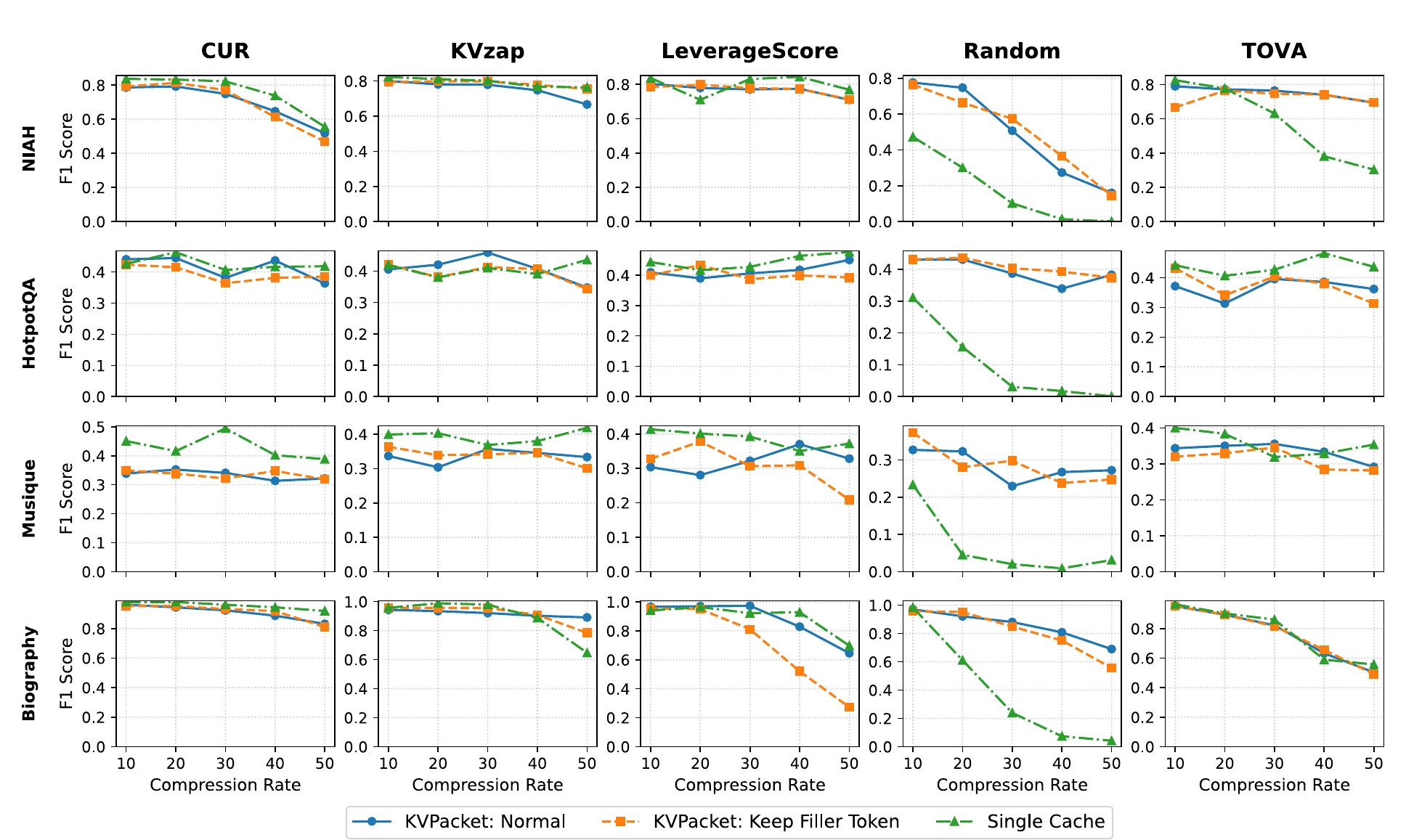}
    \caption{F1 Score vs. Compression rate of Llama-3.1-8B-Instruct model on four datasets with different compression methods}
    \label{fig:compress}
\end{figure}

Fig.~\ref{fig:compress} evaluates Llama-3.1-8B-Instruct under five compression methods (CUR \citep{curcompress}, KVzap \citep{kvzap}, LeverageScore \citep{compactor}, TOVA \citep{tova}, and random pruning) from the KVPress library \citep{devoto2025expectedattention}, at compression rates of 10\%–50\%, under three configurations: \textbf{KVPacket Normal} (compression over full wrapped cache), \textbf{KVPacket Keep Filler} (filler tokens excluded from pruning), and \textbf{Single Cache} (compression over Full Recompute cache). Notably, KV Packet proves significantly more robust than the baseline under random pruning, maintaining a much flatter performance profile. Furthermore, the KVPacket Normal setting generally outperforms the Keep Filler Token variant. This suggests that allowing the compression algorithm a more flexible pruning strategy over the entire wrapped cache is beneficial and indicates that our trained fillers are inherently resilient to KV compression.

\begin{table}[ht]
\centering
\caption{Cross-domain generalization analysis. Rows represent the dataset used to train the KV Packet adapters for Llama-3.1-8B-Instruct, while columns indicate the evaluation performance (F1 score) across different domains.}
\label{tab:cross_domain}
\begin{tabular}{@{}lcccc@{}}
\toprule
 & \multicolumn{4}{c}{\textbf{Evaluation Dataset}} \\ \cmidrule(l){2-5} 
\textbf{Training Dataset} & Biography & HotpotQA & MusiQue & NIAH \\ \midrule
No Recompute & 0.06 & 0.24 & 0.01 & 0.39 \\ \midrule
Biography               & \textbf{0.96} & 0.18 & 0.16 & 0.44 \\
HotpotQA                & 0.88 & \textbf{0.44} & 0.39 & 0.36 \\
MusiQue                 & 0.87 & 0.36 & 0.37 & 0.34 \\
Needle-in-a-Haystack (NIAH)                    & 0.54 & 0.28 & 0.07 & \textbf{0.80} \\ \midrule
Cross-domain (Mixture)     & \textbf{0.95} & \textbf{0.42} & \textbf{0.43} & 0.71 \\ \bottomrule
\end{tabular}
\end{table}

\subsection{Ablation Study}

\subsubsection{Cross-Domain Generalization and Mixture Training}
To evaluate the transferability and robustness of our learnable adapters, we conducted a cross-domain experiment in which KV Packets were trained on a specific source domain and evaluated across all other target domains. As detailed in Table~\ref{tab:cross_domain}, the No Recompute baseline fails consistently across all benchmarks ($0.01$–$0.39$), while adapters trained on specialized tasks like Biography achieve near-perfect in-domain performance ($0.96$), but struggle with the logical complexity of multi-hop reasoning tasks such as HotpotQA ($0.18$) and MusiQue ($0.16$). Conversely, the NIAH-trained adapter—while highly effective at synthetic retrieval ($0.80$)—shows limited generalization to natural-language nuances. The cross-domain adapter, trained on a diverse mixture of all four datasets, performs well across the evaluated domains. It effectively bridges the gap between specialized domains, approaching the performance of the Biography-specific model ($0.95$) and outperforming the single-domain adapters on the MusiQue reasoning benchmark ($0.43$). This suggests that heterogeneous training can teach cache-stitching patterns that transfer across related tasks. A cross-domain adapter requires representative unlabeled training traffic and validation of its task mixture; its applicability to arbitrary unseen domains is not established.

\subsubsection{Attention Dynamics with KV Packet}
\begin{figure}[t]
    \centering
    \includegraphics[width=\linewidth]{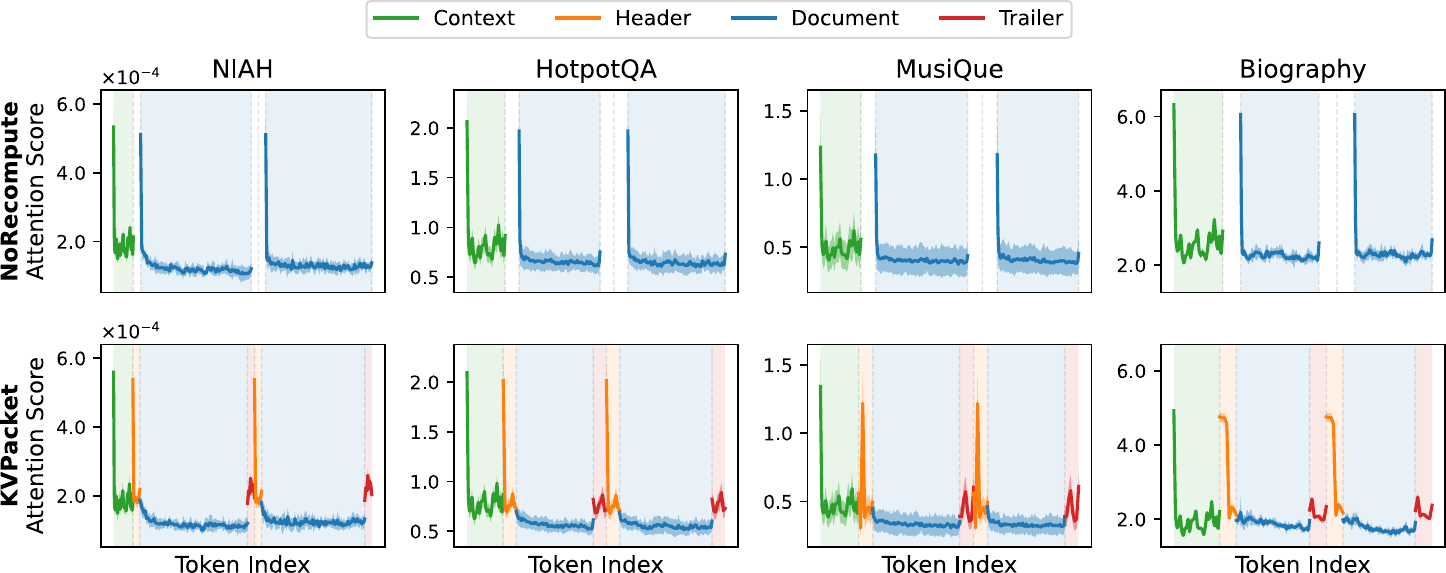}
    \caption{Query-to-context attention scores of the No Recompute and KV Packet methods. The attention scores are averaged by all layers, all query tokens, and 50 randomly selected samples from each dataset. The token positions for the No Recompute method in the figure are adjusted to align with those of the KV Packet method with filler tokens. 
    }
    \label{fig:attn_score_plot}
\end{figure}

To understand how KV Packet resolves contextual dependencies, we analyze the attention distributions of the Llama model across various datasets. We measure the averaged attention scores of a query token directed toward its preceding context, comparing the naive cache concatenation baseline (NoRecompute) with our KV Packet approach. Fig.~\ref{fig:attn_score_plot} visualizes these scores by context: the task prompt, the document tokens, and the learned adapters. Since the adapters are absent in the baseline, we manually shifted the token positions to align with the KV Packet's positions. Under naive concatenation, sharp attention spikes emerge at the start of every document block, because the model perceives each cached document's first token as a sequence-initial token, triggering the attention sink effect \citep{xiao2024efficientstreaminglanguagemodels} at mid-sequence positions and fragmenting global reasoning.

KV Packet resolves this: the Headers and Trailers absorb the sink mass that would otherwise concentrate on document tokens, as evidenced by the higher attention scores on adapter tokens than on document tokens. This redirects attention away from non-semantic boundaries and restores a smooth distribution over document content. These observations explain the strong empirical performance of even a small number of adapter tokens and strongly validate our boundary artifact hypothesis.

%% file: 06_conclusion.tex
\section{Conclusion}
\label{sec:conclusion}
We presented KV Packet, a recomputation-free framework for context-independent KV caching. By wrapping immutable document caches with lightweight trainable Headers and Trailers, we address boundary artifacts that arise from naive cache concatenation, as validated by attention score analysis. Our self-supervised distillation objective trains these adapters without human-labeled data, keeping base model weights frozen. KV Packet integrates seamlessly with existing KV compression techniques, a compatibility that recomputation-based methods fundamentally cannot match, and reduces inference-time FLOPs by 5--6 orders of magnitude while recovering much of the same base model's Full Recompute accuracy on retrieval and reasoning benchmarks.

\textbf{Limitations and Future Work}
\begin{enumerate}[leftmargin=1.2em, itemsep=0pt, topsep=2pt]
\item Adapter effectiveness assumes the retrieval corpus is reasonably aligned with the training distribution; generalization to highly out-of-distribution domains remains an open question.
\item Current evaluations cover only a limited set of model families due to the engineering complexity of validating recomputation baselines for each architecture.
\item Highly fragmented retrieval can limit absolute multi-hop accuracy. Grouping documents into larger packets mitigates this in the controlled MuSiQue experiment, but packet granularity remains an important cache-construction parameter (Appendix~\ref{app:granularity}).
\item Long-context performance remains limited: raw recovery decreases from 91.5\% at 8k/4 packets to 76.5\% at 32k/16 packets. Context length and packet count increase together, and we have not evaluated 128k contexts (Appendix~\ref{app:long_context}).
\item Long-form generation retains a quality gap relative to Full Recompute. Although ROUGE and reference-summary perplexity are close, order-balanced pairwise judgments favor Full Recompute (Appendix~\ref{app:generation}).
\end{enumerate}


%% file: 07_appendix.tex
\clearpage
\appendix\section{Appendix}

\begin{table}[ht]
\centering
\caption{Ablation study on adapter token choice. Rows represent different KV Packet configurations for Llama-3.1-8B-Instruct, including a no-adapter baseline (No Recompute), fixed-token handcrafted adapters, and the trained adapter. Columns indicate evaluation performance (F1 score) across datasets.}
\label{tab:adapter_ablation}
\begin{tabular}{@{}lcccc@{}}
\toprule
 & \multicolumn{4}{c}{\textbf{Evaluation Dataset}} \\ \cmidrule(l){2-5}
\textbf{KV Packet Configuration} & Biography & HotpotQA & MusiQue & NIAH \\ \midrule
No Recompute & 0.06 & 0.24 & 0.01 & 0.39 \\ \midrule
\texttt{<|begin\_of\_text|>} & 0.01 & 0.00 & 0.00 & 0.17 \\
\texttt{.}                   & 0.20 & 0.21 & 0.02 & 0.29 \\
\texttt{<eot\_id>}           & 0.25 & 0.22 & 0.03 & 0.33 \\
\texttt{\textbackslash n\textbackslash n} & 0.22 & \textbf{0.32} & 0.03 & 0.29 \\
\texttt{" "}                 & 0.21 & 0.24 & 0.08 & 0.28 \\
\texttt{\textbackslash n}    & \textbf{0.29} & 0.24 & 0.07 & 0.33 \\
\texttt{skip}                & 0.15 & 0.11 & \textbf{0.11} & 0.27 \\
\texttt{document}            & 0.15 & 0.12 & \textbf{0.11} & 0.26 \\ \midrule
KV Packet (trained)          & 0.94 & 0.44 & 0.38 & 0.80 \\ \bottomrule
\end{tabular}

\end{table}

\subsection{Ablation Study: Handcrafted vs.\ Trained Adapters}
To validate the importance of the proposed training objective, we compare KV Packet (trained) against a set of handcrafted adapters constructed by repeating a single fixed token from the model's vocabulary. Results are reported in Table~\ref{tab:adapter_ablation}.

Handcrafted adapters provide inconsistent improvements over the No Recompute baseline. For instance, the single escape configuration \texttt{\textbackslash n} achieves an F1 of 0.29 on Biography and 0.33 on NIAH, outperforming No Recompute (0.06 and 0.39 respectively) on the former but not on the latter. Similarly, double escape \texttt{\textbackslash n\textbackslash n} improves HotpotQA from 0.24 to 0.32, yet yields no gain on MusiQue. No single token consistently outperforms others across all four datasets, and some tokens, notably \texttt{<|begin\_of\_text|>}, severely degrade performance to near zero on HotpotQA and MusiQue. This instability suggests that the choice of delimiter token is highly dataset-sensitive and cannot be reliably determined without exhaustive search.

A deeper issue is that discrete vocabulary tokens carry inherent semantic and syntactic associations learned during pretraining. When repeated as boundary adapters, these tokens cannot function as purely structural delimiters — the model's attention mechanism responds to their semantic content in addition to their positional role, interfering with the document tokens rather than cleanly absorbing boundary artifacts. In contrast, the trained continuous adapter vectors are optimized directly to absorb attention sinks and smooth token distribution shifts at block boundaries, without any semantic baggage. This is reflected in the results: KV Packet (trained) achieves F1 scores of 0.94, 0.44, 0.38, and 0.80 on Biography, HotpotQA, MusiQue, and NIAH respectively, outperforming the best handcrafted token on every dataset by a substantial margin — for example, exceeding the best handcrafted result on Biography by 0.65 and on MusiQue by 0.27.

\subsection{Parallel Training Implementation}
\label{app:parallel_training}
\begin{figure}[!htbp]
\centering
\includegraphics[width=0.80\linewidth]{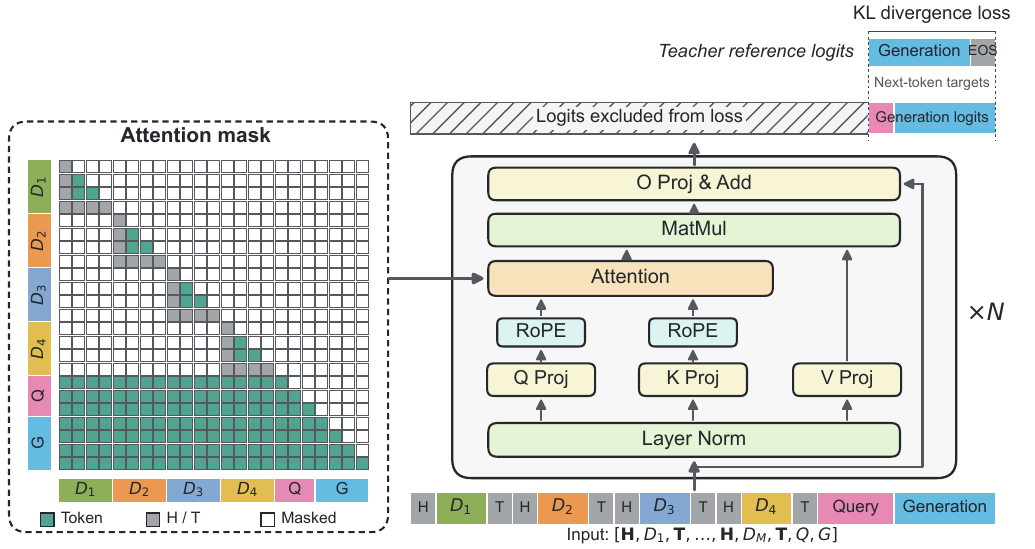}
\caption{Parallel training via block-diagonal masking. \textit{Left}: The attention mask used during training. Document blocks attend only within themselves (block-diagonal structure), while the query and generation tokens attend to all preceding tokens (full causal attention). Hatched cells indicate logits excluded from the loss computation. \textit{Right}: The transformer layer processes the complete sequence in a single forward pass. The KL divergence loss is computed only between the generation logits and the teacher reference distribution.}
\label{fig:parallel_training}
\end{figure}

The sequential training procedure described in Section~\ref{sec:methods} — which involves independent KV generation per document, concatenation, and a separate forward pass — can be unified into a single forward pass for efficient batch training. We describe this equivalent parallel implementation here.
\textbf{Block-Diagonal Attention Mask.} The key insight is that the isolated caching environment of the sequential procedure can be exactly simulated by a carefully designed attention mask. As illustrated in Fig.~\ref{fig:parallel_training}, we process the complete sequence $X_{total} = [\mathbf{H}, D_1, \mathbf{T}, \mathbf{H}, D_2, \mathbf{T}, \dots, \mathbf{H}, D_M, \mathbf{T}, Q, G] $ in a single forward pass using a block-diagonal attention mask. Specifically, the mask enforces the following attention rules:
\begin{itemize}[leftmargin=1.2em, itemsep=0pt, topsep=2pt]
    \item Each document block $[\mathbf{H}, D_i, \mathbf{T}]$ attends only within itself, simulating independent offline KV cache generation.
    \item The query $Q$ and generation tokens $G$ attend to all preceding tokens, including all document packets, consistent with the full-context assumption at inference time.
\end{itemize}

\textbf{Loss Calculation.} Only the logits corresponding to the generation tokens $G$ are used to compute the KL divergence loss against the teacher reference distribution $\mathbf{P}_{teacher}$, which is obtained from a standard causal forward pass over $X_{total}$. The hatched region in Fig.~\ref{fig:parallel_training} denotes logits that are masked out and excluded from the loss computation.

\textbf{Mathematical Equivalence.} This parallel formulation is mathematically equivalent to the sequential procedure: the block-diagonal mask ensures that the KV states of each document block are computed in isolation, identical to what would be produced by independent caching. The query and generation tokens then attend across all packets via the concatenated KV cache, replicating the inference-time composition. This equivalence means the two implementations are interchangeable, and the parallel version is adopted for training due to its significantly higher throughput.

\subsection{Training Objective}
We compare two training objectives for the adapter: KL divergence against the teacher distribution ($\mathcal{L}_{KL}$) and standard cross-entropy against the teacher labels ($\mathcal{L}_{CE}$). As shown in Table~\ref{tab:loss_ablation}, the two objectives perform comparably on retrieval tasks: on Biography, label training achieves 0.979 versus 0.943 for logits, and on NIAH, label training achieves 0.833 versus 0.803. However, on reasoning tasks that require generating longer sequences, the gap widens considerably: on MusiQue, logits training achieves 0.375 compared to only 0.226 for label training, and on HotpotQA the difference is smaller but consistent (0.436 vs. 0.430). We attribute this discrepancy to the nature of the two tasks: retrieval tasks require generating short, specific answers where the label signal is sufficient, whereas reasoning tasks require generating longer sequences where the full output distribution carries richer information. Since our objective is to align the adapter's output with the model's own full-attention behavior rather than with external ground truth, the logits-based objective better preserves the fine-grained distributional details necessary for multi-step reasoning, and is therefore adopted as the default training objective. Nevertheless, the precise mechanism underlying this difference, and whether the optimal choice of training objective varies across task types or model families, remains unclear and we leave a more thorough investigation for future work.

\begin{table}[th]
\centering
\caption{Ablation study on training objective. F1 scores are reported for KL divergence against teacher logits ($\mathcal{L}_{KL}$) and cross-entropy against teacher-generated labels ($\mathcal{L}_{CE}$) across four datasets.}
\label{tab:loss_ablation}
\begin{tabular}{@{}lcccc@{}}
\toprule
 & \multicolumn{4}{c}{\textbf{Evaluation Dataset}} \\ \cmidrule(l){2-5}
\textbf{Training Objective} & Biography & HotpotQA & MusiQue & NIAH \\ \midrule
$\mathcal{L}_{KL}$ (Logits)  & 0.943 & \textbf{0.436} & \textbf{0.375} & 0.803 \\
$\mathcal{L}_{CE}$ (Label)   & \textbf{0.979} & 0.430 & 0.226 & \textbf{0.833} \\ \bottomrule
\end{tabular}
\end{table}

\subsection{Sensitivity Analysis and Storage Overhead}
\label{app:sensitivity}
\begin{figure}[ht]
    \centering
    \includegraphics[width=\linewidth]{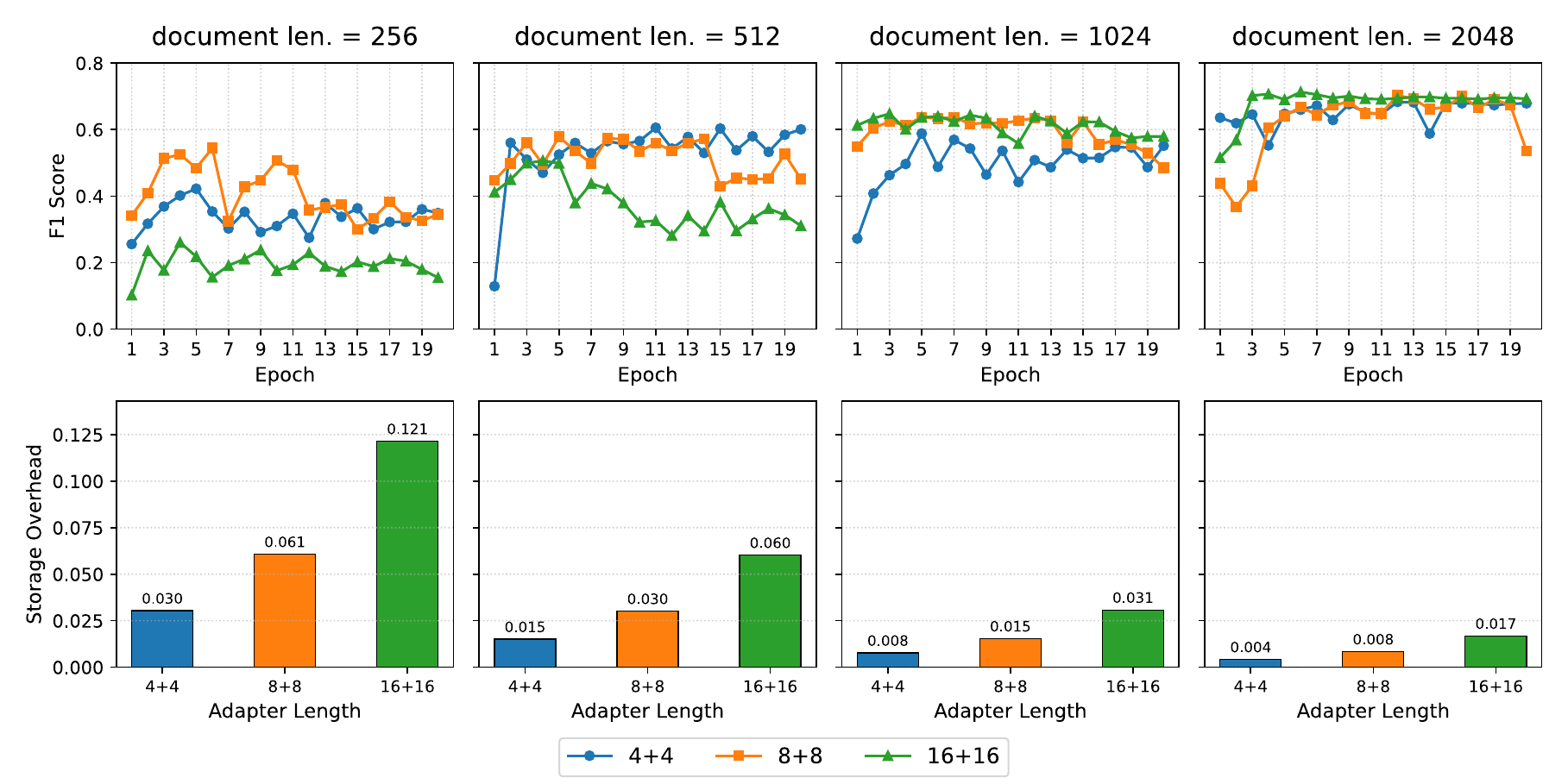}
    \caption{Sensitivity analysis of KV Packet on NIAH (Llama-3.1-8B-Instruct, 8k context). \textit{Top row}: F1 score vs.\ training epoch for each document length and adapter length configuration. \textit{Bottom row}: storage overhead introduced by each adapter length configuration at each document length.}
    \label{fig:sensitivity}
\end{figure}

We conduct a sensitivity analysis on the NIAH dataset using Llama-3.1-8B-Instruct to study the effect of document length and adapter length on KV Packet's performance and \textbf{storage overhead}. We fix the total context length at $8k$ tokens and vary the document length over \{256, 512, 1024, 2048\} tokens and the adapter length over \{4+4, 8+8, 16+16\} tokens, training each configuration for 20 epochs. Results are shown in Fig.~\ref{fig:sensitivity}.

\textbf{Convergence speed} Across nearly all configurations, the F1 score stabilizes within 3-5 epochs and shows no consistent improvement beyond that point, indicating that the 20-epoch training budget is redundant. This suggests that KV Packet adapters can be trained efficiently with a small number of gradient steps, which is favorable for practical deployment.

\textbf{Storage overhead} The storage overhead introduced by KV Packet is determined jointly by the adapter length and the document length, and remains small in realistic settings. The worst case — document length 256 with adapter length 16+16 — incurs an overhead of 12.1\%, but this configuration is also the least performant (see below). For document lengths of 512 or above, which better reflect real-world RAG scenarios, the overhead ranges from 0.4\% to 6.0\%, which is negligible relative to the KV cache storage of the documents themselves.

\textbf{Adapter length sensitivity} Longer adapters do not always yield better performance. For document lengths of 256 and 512, the 16+16 configuration performs consistently worse than 4+4 and 8+8. We attribute this to the fact that shorter documents result in a larger number of retrieved chunks within the fixed $8k$ context (up to 32 documents for 256-token chunks), and optimizing a large number of adapter tokens jointly across many boundaries makes the training objective harder to satisfy. Better hyperparameter tuning could potentially mitigate this issue, and we leave this to future work.

\textbf{Effect of document count} There is a consistent performance degradation as the number of retrieved documents increases. Across all adapter configurations, average F1 scores are lower for shorter document lengths, since the fixed $8k$ context accommodates more documents (32 for 256-token chunks versus 4 for 2048-token chunks), increasing the number of block boundaries that the adapters must bridge simultaneously. While retrieving 32 documents in a single query is uncommon in practice, we acknowledge this as a current limitation of KV Packet and leave improving performance under high document counts as a direction for future work.

\input{08_review_experiments}

%% file: 08_review_experiments.tex
\subsection{Direct Comparison with KVLink}
\label{app:kvlink}
We compare stock Llama-3.2-3B with KV Packet against the released Llama3B-KVLink5 checkpoint \citep{kvlink}. Both methods use the same 100 seed-42 examples per dataset, a shared prompt, greedy decoding, BF16 eager attention, and one A100 80~GB. KVLink fine-tunes the backbone, whereas KV Packet keeps it frozen, so their Full Recompute references differ. Table~\ref{tab:kvlink_accuracy} reports absolute F1 together with raw recovery, defined as the cache-reuse F1 divided by Full Recompute F1 for the same checkpoint. Recovery is higher for KV Packet on all four datasets; this is a comparison of accuracy preservation, not an assumption that the checkpoints have identical capabilities.

\begin{table}[htbp]
\centering
\caption{KVLink comparison. Full denotes recomputation using each method's own checkpoint. Recovery is reported as a percentage; the mean averages the four dataset recovery values.}
\label{tab:kvlink_accuracy}
\begin{tabular}{@{}lrrrrrr@{}}
\toprule
 & \multicolumn{3}{c}{KV Packet (stock)} & \multicolumn{3}{c}{KVLink-5 (fine-tuned)} \\
\cmidrule(lr){2-4}\cmidrule(l){5-7}
Dataset & F1 & Full & Recovery & F1 & Full & Recovery \\ \midrule
Biography & 0.8044 & 0.8811 & 91.3\% & 0.8811 & 0.9744 & 90.4\% \\
NIAH & 0.7474 & 0.8088 & 92.4\% & 0.6589 & 0.7628 & 86.4\% \\
HotpotQA & 0.3336 & 0.4289 & 77.8\% & 0.1320 & 0.2977 & 44.3\% \\
MuSiQue & 0.1340 & 0.2797 & 47.9\% & 0.0702 & 0.2727 & 25.7\% \\
Mean & -- & -- & 77.4\% & -- & -- & 61.7\% \\
\bottomrule
\end{tabular}
\end{table}

For serving efficiency, the common preamble and document caches are prepared offline for both methods. KVLink's links attend to preceding segments, so their context-dependent states must be computed online. We evaluate both sequential KVLink and a fused implementation, verifying first-token equivalence on all evaluated examples. Table~\ref{tab:kvlink_efficiency} reports total online FLOPs, including query forward computation; these totals differ from the cache-preparation-only FLOPs in Fig.~\ref{fig:llama_result}. KV Packet uses 44.2\% fewer online FLOPs on HotpotQA and 57.9\% fewer on MuSiQue than fused KVLink. Fused TTFT is comparable, with no consistent latency advantage for either method.

\begin{table}[htbp]
\centering
\caption{KVLink serving comparison. TTFT is in milliseconds and online FLOPs in trillions ($10^{12}$), including query forward computation.}
\label{tab:kvlink_efficiency}
\begin{tabular}{@{}llrr@{}}
\toprule
Dataset & Method & TTFT (ms) & Online FLOPs (T) \\ \midrule
HotpotQA & KV Packet & 90.03 & 0.3900 \\
HotpotQA & KVLink sequential & 340.09 & 0.6870 \\
HotpotQA & KVLink fused & 93.73 & 0.6983 \\ \midrule
MuSiQue & KV Packet & 173.03 & 0.4732 \\
MuSiQue & KVLink sequential & 635.10 & 1.0837 \\
MuSiQue & KVLink fused & 167.85 & 1.1252 \\
\bottomrule
\end{tabular}
\end{table}

The methods also differ in supervision and placement: KVLink uses labeled or GPT-4-generated targets and trailer-side link tokens, while KV Packet self-distills the frozen model's logits with Header and Trailer adapters. Consequently, this comparison does not isolate the effect of placement, supervision, or backbone treatment individually. We have not run a decisive header-versus-trailer ablation. Block-Attention and CacheClip were not reproduced in this comparison: we prioritized KVLink as the closest learning-based baseline with released checkpoints, rather than claiming coverage of every related method.

\subsection{Expanded Mechanistic Decomposition}
\label{app:mechanism}
We expand the HotpotQA decomposition to 2,000 unique examples using Llama-3.1-8B-Instruct, with matched examples across four conditions. No Recompute concatenates isolated document caches without adapters; KV Packet adds trained H/T while retaining isolated document encoding; the hybrid restores full cross-document attention with the trained H/T still present; Full Recompute uses ordinary full attention without H/T. Table~\ref{tab:mechanism_scores} reports pooled token-count Micro F1, example-averaged Macro F1, and exact match (EM).

\begin{table}[htbp]
\centering
\caption{HotpotQA decomposition on 2,000 unique examples using Llama-3.1-8B-Instruct. The hybrid retains H/T and restores full cross-document attention.}
\label{tab:mechanism_scores}
\begin{tabular}{@{}lrrr@{}}
\toprule
Condition & Micro F1 & Macro F1 & EM \\ \midrule
No Recompute & 0.207 & 0.215 & 0.156 \\
KV Packet & 0.382 & 0.423 & 0.309 \\
Full Attention + H/T & 0.428 & 0.443 & 0.328 \\
Full Recompute & 0.408 & 0.460 & 0.344 \\
\bottomrule
\end{tabular}
\end{table}

We compute paired-bootstrap 95\% confidence intervals for the Macro F1 differences between matched conditions (Table~\ref{tab:mechanism_ci}). KV Packet improves Macro F1 over No Recompute by 0.2089 and recovers 84.9\% of the Full-Recompute-over-No-Recompute improvement. Here, recovered improvement is $(F_{\mathrm{KV}}-F_{\mathrm{No}})/(F_{\mathrm{Full}}-F_{\mathrm{No}})$, distinct from raw recovery $F_{\mathrm{KV}}/F_{\mathrm{Full}}$. Restoring cross-document attention adds 0.0198 Macro F1, with a positive confidence interval. Boundary repair therefore accounts for most of the recovered performance in this experiment, while cross-document attention has a smaller, statistically resolved positive contribution. Estimates are reported separately from the rounded scores in Table~\ref{tab:mechanism_scores}.

\begin{table}[htbp]
\centering
\caption{Paired-bootstrap uncertainty for Macro F1 differences in the expanded HotpotQA decomposition.}
\label{tab:mechanism_ci}
\begin{tabular}{@{}lrc@{}}
\toprule
Comparison & Estimate & 95\% CI \\ \midrule
KV Packet $-$ No Recompute & +0.2089 & $[0.1888,\ 0.2293]$ \\
Full Attention + H/T $-$ KV Packet & +0.0198 & $[0.0035,\ 0.0362]$ \\
Full Recompute $-$ Full Attention + H/T & +0.0173 & $[0.0013,\ 0.0340]$ \\
\bottomrule
\end{tabular}
\end{table}

To quantify evaluation-set variability at a 100-example test size, we partition the 2,000-example pool into 20 non-overlapping subsets. Mean raw recovery across these subsets is $92.5\%\pm10.4$ percentage points (mean $\pm$ sample standard deviation), with a reported 95\% confidence interval of $[88.2\%,97.1\%]$ for mean recovery. This subset analysis measures evaluation-set variability; it is distinct from independent adapter training runs and from the paired-bootstrap differences above.

\subsection{Long-Context Scaling}
\label{app:long_context}
On NIAH, we reuse an 8k-trained Llama-3.1-8B adapter without retraining, keep packet length near 2k tokens, and increase packet count from four to sixteen. Each point is evaluated on 100 frozen examples (Table~\ref{tab:long_context}). Raw F1 recovery declines from 91.5\% at 8k to 76.5\% at 32k, although KV Packet remains above No Recompute. At 32k, it recovers 60.1\% of the Full-over-No improvement. Context length and packet count increase together, so this experiment does not isolate their individual effects. We have not trained an adapter at 32k or evaluated 128k contexts.

\begin{table}[htbp]
\centering
\caption{NIAH long-context evaluation with an unchanged 8k-trained Llama-3.1-8B adapter. Recovery is raw F1 recovery relative to Full Recompute, as reported.}
\label{tab:long_context}
\begin{tabular}{@{}lrrrr@{}}
\toprule
Context / packets & Full Recompute & No Recompute & KV Packet & Recovery \\ \midrule
8k / 4 & 0.868 & 0.613 & 0.795 & 91.5\% \\
16k / 8 & 0.845 & 0.453 & 0.742 & 87.9\% \\
32k / 16 & 0.801 & 0.328 & 0.612 & 76.5\% \\
\bottomrule
\end{tabular}
\end{table}

\subsection{Long-Form Generation Quality}
\label{app:generation}
We train a task-specific Llama-3.1-8B adapter on 256 self-generated Multi-News examples \citep{fabbri2019multinews}, select epoch five using a disjoint calibration set, and evaluate 100 frozen validation examples with greedy generation capped at 384 tokens. Table~\ref{tab:generation} reports reference-based ROUGE, gold-summary perplexity, and repeated trigrams. Gold-summary perplexity changes from 6.592 to 6.603 (+0.18\%). Paired-bootstrap intervals for all three ROUGE differences include zero; the ROUGE-1 interval for KV Packet minus Full Recompute is $[-0.0105,0.0068]$. Mean output length is 198.5 words for Full Recompute and 201.4 for KV Packet, and neither method reaches the generation cap.

\begin{table}[htbp]
\centering
\caption{Multi-News generation evaluation. ROUGE values are reported on a 0--1 scale; gold PPL is reference-summary perplexity.}
\label{tab:generation}
\begin{tabular}{@{}lrrrrr@{}}
\toprule
Method & ROUGE-1 & ROUGE-2 & ROUGE-L & Gold PPL & Repeated trigrams \\ \midrule
Full Recompute & 0.4284 & 0.1344 & 0.2041 & 6.592 & 1.64\% \\
KV Packet & 0.4266 & 0.1294 & 0.1999 & 6.603 & 2.03\% \\
\bottomrule
\end{tabular}
\end{table}

A blinded Qwen3-32B pairwise judge assesses factuality, coverage, integration, coherence, and fluency. To mitigate observed A/B position bias, each pair is judged in both presentation orders. Of the resulting 200 judgments for 100 example pairs, 133 favor Full Recompute (66.5\%), 63 favor KV Packet (31.5\%), and four are ties. These are 200 order-balanced judgments, not 200 distinct examples. Full Recompute generations have perplexity 1.251 under Full Recompute and 1.402 under KV Packet, showing a distributional difference in actual outputs that gold-summary perplexity does not capture. Thus, close reference-based metrics do not eliminate the residual long-form generation-quality gap. We have not tested whether a larger training budget closes it.

\subsection{MuSiQue Packet Granularity}
\label{app:granularity}
On Llama-3.2-3B, we use all 20 MuSiQue source documents and the same 100 seed-42 examples, prompts, and greedy decoder. We cross the original multi-hop question with a direct-final-hop control, and one document per packet with groups of at most four consecutive documents per packet (Max4). Grouping preserves the concatenated document-token stream, making Full Recompute identical across packetizations. Table~\ref{tab:granularity_crossdomain} evaluates one cross-domain adapter trained across Biography, HotpotQA, NIAH, GSM8K \citep{cobbe2021gsm8k}, and all four MuSiQue configurations. This is a separate 3B configuration from the original four-task Llama-3.1-8B mixture in Table~\ref{tab:cross_domain}.

\begin{table}[htbp]
\centering
\caption{MuSiQue Macro F1 for the cross-domain Llama-3.2-3B adapter, with fixed retrieved content. Max4 groups up to four consecutive documents. Recovery percentages are reported alongside rounded F1 scores.}
\label{tab:granularity_crossdomain}
\begin{tabular}{@{}llrrrr@{}}
\toprule
Packing & Question & No Recompute & KV Packet & Full Recompute & Recovery \\ \midrule
One/doc & Direct & 0.040 & 0.316 & 0.449 & 70.4\% \\
One/doc & Multi-hop & 0.036 & 0.133 & 0.235 & 56.7\% \\
Max4 & Direct & 0.108 & 0.414 & 0.449 & 92.2\% \\
Max4 & Multi-hop & 0.001 & 0.208 & 0.235 & 88.5\% \\
\bottomrule
\end{tabular}
\end{table}

Grouping raises cross-domain multi-hop raw recovery by 31.8 percentage points, from 56.7\% to 88.5\%; under Max4, the direct-versus-multi-hop recovery gap is 3.7 points. A task-specific adapter reproduces the improvement (Table~\ref{tab:granularity_task}). Recovery above 100\% is a finite-sample point estimate at $n=100$, rather than evidence of superior underlying capability. These controlled results support packet fragmentation as a major factor in this setting. Larger packets reduce isolated-encoding boundaries and allow attention within the grouped documents, so the experiment does not establish that reasoning depth is irrelevant for all workloads. Packet granularity is a cache-construction parameter: any grouped packet must be built as that unit before reuse.

\begin{table}[htbp]
\centering
\caption{MuSiQue Macro F1 with a task-specific adapter under the same packet-granularity protocol.}
\label{tab:granularity_task}
\begin{tabular}{@{}llrrrr@{}}
\toprule
Packing & Question & No Recompute & KV Packet & Full Recompute & Recovery \\ \midrule
One/doc & Direct & 0.040 & 0.352 & 0.449 & 78.3\% \\
One/doc & Multi-hop & 0.036 & 0.118 & 0.235 & 50.4\% \\
Max4 & Direct & 0.108 & 0.475 & 0.449 & 105.8\% \\
Max4 & Multi-hop & 0.001 & 0.250 & 0.235 & 106.5\% \\
\bottomrule
\end{tabular}
\end{table}